\documentclass[Royal,times,sagev,doublespace]{sagej}
\usepackage{moreverb,url}

\usepackage[colorlinks,bookmarksopen,bookmarksnumbered,citecolor=red,urlcolor=red]{hyperref}

\usepackage{graphicx}
\usepackage{amsmath}
\usepackage{bm}

\newcommand\BibTeX{{\rmfamily B\kern-.05em \textsc{i\kern-.025em b}\kern-.08em
T\kern-.1667em\lower.7ex\hbox{E}\kern-.125emX}}

\newcommand{\hv}{H}

\newcommand{\tsetsize}{m}

\newcommand{\predfun}{f}

\newcommand{\ivector}{x}
\newcommand{\wvector}{w}

\newcommand{\threshold}{t}

\begin{document}

\runninghead{}

\title{Tournament Leave-pair-out Cross-validation for Receiver Operating Characteristic (ROC) Analysis}

\author{Ileana Montoya Perez\affilnum{1,2}, Antti Airola\affilnum{1}, Peter J. Bostr\"{o}m\affilnum{2}, Ivan Jambor\affilnum{3,4} and Tapio Pahikkala\affilnum{1}}

\affiliation{\affilnum{1}Department of Future Technologies, University of Turku, Turku, Finland\\
\affilnum{2}Department of Urology, Turku University Hospital, Turku, Finland\\
\affilnum{3}Department of Diagnostic Radiology, Turku University Hospital, Turku, Finland\\
\affilnum{4}Department of Radiology, Icahna School of Medicine at Mount Saina, New York, USA}

\begin{abstract} Receiver operating characteristic (ROC) analysis is widely used for evaluating diagnostic systems. Recent studies have shown that estimating an area under ROC curve (AUC) with standard cross-validation methods suffers from a large bias. The leave-pair-out (LPO) cross-validation has been shown to correct this bias. However, while LPO produces an almost unbiased estimate of AUC, it does not provide a ranking of the data needed for plotting and analyzing the ROC curve. In this study, we propose a new method called tournament leave-pair-out (TLPO) cross-validation. This method extends LPO by creating a tournament from pair comparisons to produce a ranking for the data. TLPO preserves the advantage of LPO for estimating AUC, while it also allows performing ROC analyses. We have shown using both synthetic and real world data that TLPO is as reliable as LPO for AUC estimation, and confirmed the bias in leave-one-out cross-validation on low-dimensional data. As a case study on ROC analysis, we also evaluate how reliably sensitivity and specificity can be estimated from TLPO ROC curves. \end{abstract}

\maketitle

\section{Introduction}
Diagnostic systems, such as binary classifiers, are used to predict outcomes that support decision making. In medicine, these systems help prognosis by predicting outcomes such as healthy or disease, follow-up treatment response or anticipated relapse. Diagnostic systems are usually built from data that combine events, test results and variables with the objective of discriminating between two alternatives. Often the quality of data used to generate a system is uncertain, affecting its accuracy. Hence, a good assessment of the system degree of accuracy is crucial.\citep{swets1998measuring} 

To evaluate the discrimination ability of a binary classifier, the receiver operating characteristic (ROC) analysis is a popular approach. It allows visualizing, comparing and selecting classifiers based on their performance. The ROC curve depicts the performance of a classifier across various decision thresholds, while the area under the ROC curve (AUC) quantifies the classification error. The AUC value can be interpreted as the probability of the classifier ranking a randomly chosen positive unit (e.g. diseased subject or case) higher than a randomly chosen negative unit (e.g. healthy subject or control).\cite{hanley82auc} In contrast to many other performance measures, AUC is invariant to skewed class distribution and unequal classification error costs.\cite{fawcett2006introduction}

In medical studies, classifiers are usually obtained from data sets where variables are measured from relatively small numbers of sample units and the classes are often highly imbalanced. These data sets are challenging when training and testing classifiers and ROC analysis must be used with caution in this scenario.\cite{hanczar2010small} 
Ideally, the performance of a classifier should be evaluated on independent data (i.e. data not used for training the classifier). In practice, large enough independent data may not be available or cannot be spared when building the classifier. Therefore, in many cases, cross-validation methods such as leave-one-out (LOO) and K-fold are used to estimate the performance of a classifier. However, several experimental studies have shown that LOO and many other cross-validation methods are biased for AUC estimation. This bias is caused by the pooling procedure, where predictions from different rounds of cross-validation are pooled together in order to compute the ROC curve and AUC. The pooling thus violates the basic assumption that the predictions are made by a single classifier, often leading to systematic biases.\cite{parker2007stratification,forman2010apples,airola2011experimental,perlich2011cross,smith2014correcting} As an alternative, a leave-pair-out (LPO) cross-validation that results in an almost unbiased AUC estimation was proposed and tested.\cite{airola2011experimental} Moreover, a study that used real-world clinical data sets also examined LPO and confirmed it being a reliable cross-validation method for estimating AUC.\cite{smith2014correcting} However, LPO only produces AUC estimate without providing the ranking of the data needed for performing full ROC analysis.

In this study we propose a variant of LPO cross-validation, the tournament leave-pair-out (TLPO) cross-validation. TLPO constructs a tournament from paired comparisons obtained by carrying out LPO cross-validation over all sample unit pairs. The ROC analysis can be then subsequently carried out on the scores determined by the tournament.\cite{kendall1940method,harary1966theory} In the literature, it is shown that such tournament scores are guaranteed to produce a good ranking for the data (see e.g. \cite{Coppersmith2010ordering} for a formal analysis and proof). Furthermore, through experiments on both synthetic and real medical data, we evaluate LOO, LPO, and TLPO AUC estimates from two well established classification methods: ridge regression and k-nearest neighbors (KNN). The experimental results show that the TLPO is as reliable as LPO for estimating AUC, while enabling full ROC analysis.
\section{Preliminaries}
ROC analysis is commonly used to assess the accuracy of classifiers that produce real-valued outputs. We assume a set of $\tsetsize$ sample units, divided into the so-called positive and negative classes. In a typical application, the sample units would correspond to patients, and the classes to absence or presence of a certain disease. We denote by $\mathcal{I}=\{1,2,...,\tsetsize\}$ the index set of these sample units, and by $\mathcal{I}_+\subset\mathcal{I}$ and $\mathcal{I}_-\subset\mathcal{I}$ the indices of the positive and the negative sample units, respectively. Note that we refer to the sample units only by their indices $i\in\mathcal{I}$, since their other properties, such as possible feature representations, are irrelevant when studying cross-validation techniques.

Let $\predfun: \mathcal{I}\rightarrow\mathbb{R}$ denote a prediction function, that maps each sample unit to a real-valued prediction indicating how likely they are to belong to the positive class. By sorting the predicted values $\predfun(1), \predfun(2),\ldots,\predfun(\tsetsize)$ the sample units may then be ordered from the one predicted most likely to belong to negative class, to the one predicted most likely to belong to positive class. In order to transform the predictions into binary classes, a threshold $\threshold$ may be set so that the sample units with smaller predictions are classified as negatives, and higher as positives. This can be described as a classifier
\begin{align*}
 C_\threshold(i) = \begin{cases} 1 & if \ \  \predfun(i)> \threshold \\
 0 &  otherwise\end{cases}.
\end{align*}
The classification performance of $C_\threshold$ is often evaluated by measuring the true positive rate (TPR), also known as sensitivity or probability of detection, and the false positive rate (FPR) as $\threshold$ is varied. Formally, these are defined as the probabilities of a positive unit getting correctly identified as positive and a non-positive unit getting wrongly identified as positive, that is
\begin{equation}\label{tpr}
\textnormal{TPR}=\frac{1}{\arrowvert\mathcal{I}_+\arrowvert}\sum_{i\in\mathcal{I}_+}C_\threshold(i)\phantom{W}\textnormal{ and }
\phantom{W}
\textnormal{FPR}=\frac{1}{\arrowvert\mathcal{I}_-\arrowvert}\sum_{i\in\mathcal{I}_-}C_\threshold(i)\;.
\end{equation}
These can be considered either as empirical estimates on finite samples or as the actual probabilities on the population level, e.g. on all future observations by letting $\tsetsize\to\infty$.

The ROC curve plots the TPR versus FPR of a classifier for all possible values of $\threshold$. A curve that represents a perfect classifier is the one with a right angle at (0, 1), which means that there is a $\threshold$ that perfectly separates positive units from negative ones. Likewise, a classifier that makes random predictions is represented by a diagonal line from (0, 0) to (1, 1). The area under the curve or AUC is the metric that quantifies the performance of the classifier independently of $\threshold$. A perfect classifier has AUC of 1.0, while a classifier that makes random predictions or predicts a constant value has AUC of 0.5.

There are different approaches for computing the AUC.\cite{lasko2005use,hanczar2010small,hanley82auc} A common approach is to plot the ROC curve by connecting the points (TPR, FPR) with straight lines and then estimating the AUC using the trapezoid rule. An equivalent way is through Wilcoxon-Mann-Whitney (WMW) statistic, \cite{bamber1975area} which consists of making all possible comparisons between pairs of positive and negative units and scoring each comparison according to the Heaviside step function.
Then, the WMW statistic can be computed on a finite sample $\mathcal{I}$ as: 
\begin{equation}\label{auc}
\hat A(\predfun)=\frac{1}{\arrowvert\mathcal{I}_{+}\arrowvert\arrowvert\mathcal{I}_{-}\arrowvert}
\sum_{i\in\mathcal{I}_{+}}\sum_{j\in\mathcal{I}_{-}}
\hv\left(\predfun(i) - \predfun(j)\right)\;,
\end{equation}
where
\begin{align*}
 H(a) = \left \{ 
\begin{array}
{r@{,\quad}l}
1 & if \ \  a > 0 \\ 0.5 & if \ \  a = 0 \\ 0 & if \ \  a < 0 
\end{array} \right.\;.
\end{align*}
is the Heaviside step function.
Again its limit:
\[
A(\predfun)=\lim_{\tsetsize\to\infty}\hat A(\predfun)\;
\]
is the true AUC of $\predfun$ also covering all future observations.

Performing ROC analysis of machine learning based classifiers is simple when having access to large amounts of data. The prediction function is learned from a training set, and the ROC curve and AUC are computed on an independent test set. Usually in small sample settings, a separate test set cannot be afforded. Testing a prediction function directly on the same data it was learned from (i.e. resubstitution) leads to highly overoptimistic results. 
Rather, methods such as bootstrapping and cross-validation are used in order to provide reliable performance estimates in small sample settings.

Bootstrapping methods, such as the one described by Harrell et al.,\citep{harrell1996multivariable} or the .632\cite{efron1983estimating} and .632+\cite{efron1997improvements} estimates allow adjusting the optimistic bias present in the resubstitution estimate. It consists of estimating the amount of bias by using a large number of bootstrap samples drawn with replacement from the original sample, and correcting the resubstitution estimate accordingly. Bootstrap methods may be expected to provide reliable performance estimates when using classical statistical approaches, such as generalized linear models, with modest amount of features. However, they are known to be vulnerable to overoptimism when high dimensional data or complex non-linear models are used. This is because they are partly based on the resubstitution estimate, which can become arbitrarily biased when using flexible enough models that can always be fitted to predict their own training data almost perfectly, such as the KNN method.
Previously, Kohavi\cite{Kohavi1995cv} and Smith et al.\cite{smith2014correcting} have experimentally confirmed the optimistic bias of boostrap compared to cross-validation. This is further confirmed in our study by experimental results presented in the supplementary materials.

Cross-validation involves splitting the available data repeatedly into two non-overlapping parts, training and test set. The training set is used to train or build the classifier and the test set to evaluate its performance. In K-fold cross-validation, the data is split in K mutually disjoint parts (i.e. folds) of equal size. Then, in turns, each fold is held out as test data while the rest of folds (K-1) are used to train a classifier for performing predictions on the test data.
In the so-called pooled K-fold cross-validation, the predictions for all the folds are combined together, and the AUC is then computed using the combined set of predictions. In averaged K-fold cross-validation, a separate AUC is computed for each test fold, and the final AUC is the average of these fold-wise estimates. While a full ROC analysis is possible in pooled K-fold, the averaged K-fold only provides an AUC estimate. A disadvantage shared by both the pooled and averaged K-fold is that with large fold sizes they are negatively biased, because a substantial part of the training set is left out in each round of cross-validation.

In the case of leave-one-out cross-validation (LOO), each unit constitutes its own fold, and the AUC estimate is calculated using the pooling approach. Formally, the AUC estimated by LOO is:
\begin{align*}
\hat A_{\textnormal{LOO}}(\predfun)=\frac{1}{\arrowvert\mathcal{I}_{+}\arrowvert\arrowvert\mathcal{I}_{-}\arrowvert}
\sum_{i\in\mathcal{I}_{+}}\sum_{j\in\mathcal{I}_{-}}
\hv\left(\predfun_{\mathcal{I}\setminus\{i\}}(i) - \predfun_{\mathcal{I}\setminus\{j\}}(j)\right)\;,
\end{align*}
where \begin{math} \predfun_{\mathcal{I}\setminus\{i\}}\end{math} and \begin{math} \predfun_{\mathcal{I}\setminus\{j\}}\end{math} are prediction functions trained without the $i$th and $j$th sample units, respectively. 

In the pooling approach, the predictions for the $i$th and the $j$th sample units  may originate from different prediction functions. This may produce biased AUC estimates with unstable learning algorithms\textemdash the ones whose predictions functions undergo major changes in response to small changes in the training data. 

Many learning algorithms produce prediction functions that can be decomposed into two components, that is $f(i)=g(i)+c$, where the first depends on the unit $i$ and the second that is independent of it. 
In the context of linear models, the prediction function often has a constant term referred to as the intercept. These constant term may bias the pooling AUC estimate. This problem is particularly severe for LOO. For example, if the training algorithm infers from the data a constant valued prediction function $f(i)=1/{p_i}-1/{n_i}$ consisting of the difference between the inverse frequencies of positive $p_i$ and negative $n_i$ units in the training set during the $i$th round of LOO, then the LOO predictions for positive units will all have a larger predictions than the negative ones, resulting to AUC value 1, even though the constant functions are not of any use for prediction.

An analogous negative bias can also emerge, when the learning algorithms tend to produce prediction functions whose values correlate with the class proportions.\cite{perlich2011cross}  In addition, experiments performed on synthetic and real data sets have shown that both pooled K-fold and LOO cross-validation estimates suffer from a high negative bias when used for AUC estimation. \cite{airola2011experimental,parker2007stratification,smith2014correcting}  Hence, using pooling for AUC estimation is very risky as it may produce arbitrarily badly biased results. 

When using averaging to estimate the AUC, as in averaged K-fold cross-validation, the negative bias caused by pooling disappears.\cite{parker2007stratification} However, it has been shown that averaging leads to high variance in the AUC estimates when using small data sets.\cite{airola2011experimental}  Another issue in K-fold, is that the value of K is constrained by the number of units in the minority class. For example, if there are more folds than positives units, the AUC for the folds without positives cannot be calculated affecting the averaged AUC. 

For a more reliable AUC estimate, LPO cross-validation has been proposed.\cite{airola2011experimental} This cross-validation method combines the strengths of pooling and averaging approaches. In LPO each positive-negative pair is held as test data and the cross-validation AUC is computed by averaging over all these pairs’ predictions, as in equation \eqref{auc}. This ensures that only pairs from the same round of cross-validation are compared, while it makes maximal use of the available training data. Formally, the LPO cross-validation estimate is defined as
\begin{align*}
\hat A_{LPO}(\predfun)=\frac{1}{\arrowvert\mathcal{I}_{+}\arrowvert\arrowvert\mathcal{I}_{-}\arrowvert}
\sum_{i\in\mathcal{I}_{+}}\sum_{j\in\mathcal{I}_{-}}
\hv\left(\predfun_{\mathcal{I}\setminus\{i,j\}}(i) - \predfun_{\mathcal{I}\setminus\{i,j\}}(j)\right)\;,
\end{align*}
where \begin{math} \predfun_{\mathcal{I}\setminus\{i,j\}}\end{math} is the prediction function trained without the $i$-th and $j$-th sample units.

\section{TLPO cross-validation}

In order to perform ROC analysis, we need a predicted ranking for the data set where the sample units ranked higher are considered more likely to belong to the positive class. As indicated previously, LPO cross-validation produces an almost unbiased AUC estimate, but it does not provide such ranking. In this section, we describe the proposed TLPO cross-validation method, a variant of LPO that applies pair comparison method \cite{kendall1940method} and round robin tournament theory \cite{harary1966theory} to produce a ranking over the data set. 

A tournament is a complete asymmetric directed graph, that is, a graph containing exactly one directed edge between each pair of vertices. In TLPO, we consider a round robin tournament in which the vertices correspond to sample units and the directions of the edges are obtained from a complete LPO cross-validation\textemdash where every possible pair of sample units is held out as test data at the time, including those pairs that belong to the same class. The edge connecting sample units $i$ and $j$ goes from the former to the latter if $\predfun_{\mathcal{I}\setminus\{i,j\}}(i)>\predfun_{\mathcal{I}\setminus\{i,j\}}(j)$, that is, its direction is determined by the order of predictions performed during a train-test split with test set $\{i,j\}$.

Given the tournament graph, the tournament score $S(i)$ for the $i$th sample unit is computed by counting the number of outgoing edges (i.e. out-degree) in the tournament graph starting from the corresponding vertex:
\begin{align*}
S(i)=\sum_{j=1}^m \hv\left(\predfun_{\mathcal{I}\setminus\{i,j\}}(i)-\predfun_{\mathcal{I}\setminus\{i,j\}}(j)\right)\;,
\end{align*}
the TLPO AUC estimate can then be computed from the tournament scores, for example, by using equation (\ref{auc})
\begin{align*}
\hat A_{TLPO}(f)=\hat A(S)\;.
\end{align*}

\begin{figure}
\centering
\includegraphics[scale=0.5]{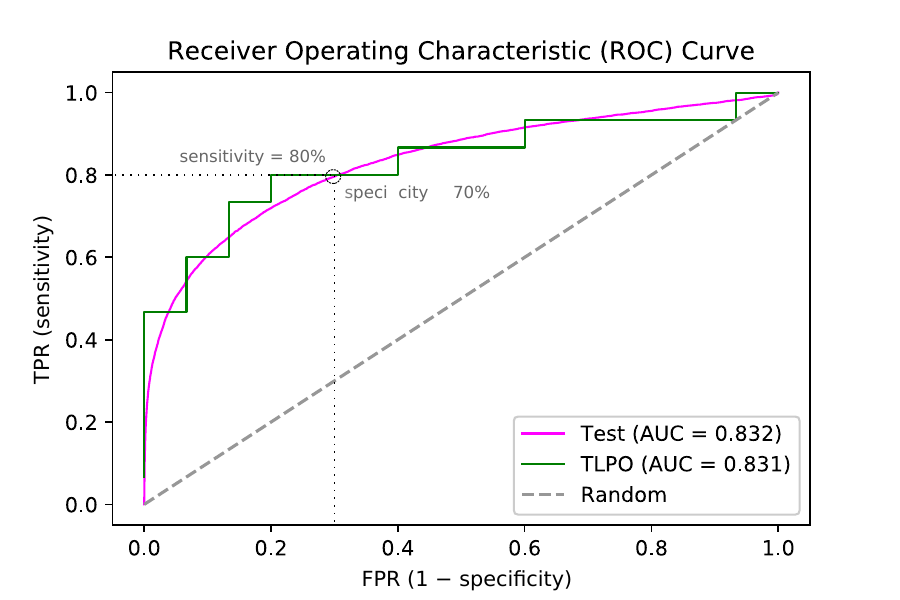}
\caption{\label{TLPO_ROC}  Example of ROC curves of a classifier evaluated by tournament cross-validation (TLPO) and by a large test data set (Test). The TLPO curve was obtained from 30 random sample units (15 positives and 15 negatives) and the rest of the data was used for the Test curve. The real medical data set and ridge regression were used.}
\end{figure}

The TLPO ranking is generated by ordering the sample units according to their scores or number of wins.
It has been shown in the literature concerning tournaments  that the above considered tournament scores provide a good ranking of the data. For the theoretical results backing this claim, we refer to \cite{Coppersmith2010ordering,balcan2008robust}. This ranking can then be used for ROC analysis to evaluate the classifier performance, as described in the previous section. In Figure \ref{TLPO_ROC}, we present an example of a ROC curve obtained using TLPO cross-validation on a randomly selected sample of 30 units (15 positives and 15 negatives) from the real medical data set (described in section 4.3) and a ROC curve using the rest of this data set as test data. The classification method used in this example was ridge regression. 

We next analyze the TLPO and compare it to the ordinary LPO. It is said that a tournament is consistent if the corresponding graph is acyclic. Otherwise, it is inconsistent, indicating that there are at least one circular triad (i.e. cyclic triple) of sample units $h$, $i$ and $j$ such that
\begin{align*}
\predfun_{\mathcal{I}\setminus\{h,i\}}(h)&<\predfun_{\mathcal{I}\setminus\{h,i\}}(i)\\ \predfun_{\mathcal{I}\setminus\{i,j\}}(i)&<\predfun_{\mathcal{I}\setminus\{i,j\}}(j)\\
\predfun_{\mathcal{I}\setminus\{h,j\}}(h)&>\predfun_{\mathcal{I}\setminus\{h,j\}}(j)\;.
\end{align*}

In TLPO cross-validation, this inconsistency rises when the learning algorithm is unstable on the sample. From the three above cases, we can see that the training data sets differ from each other only by a single sample; however, this is enough to make the three learned prediction functions so different from each other that a circular triad emerges. 

The level of inconsistency can be measured by counting the number of circular triads in the tournament graph, as explained in \cite{kendall1940method,harary1966theory,gass1998tournaments}. Based on the number of circular triads, a coefficient of consistency ($\xi$) was proposed by Kendall and Babington Smith \cite{kendall1940method}. This coefficient takes a value between 1 and 0. If $\xi = 1$ then the tournament has no circular triads; as the number of circular triads increases $\xi$ tends to zero. If $\xi = 0$ the tournament has as many  circular triads as possible. The equations for computing number of circular triads and coefficients of consistency are provided in the supplemental material.

A typical example of a perfectly stable learning algorithm, which produces a consistent tournament, is the one that always outputs the same real valued prediction function, that is, $\predfun_{\mathcal{I}\setminus\{h,i\}}=\predfun_{\mathcal{I}\setminus\{i,j\}}$, for any $h,i,j\in\mathcal{I}$. In this case, properties of the well-known ordinary WMW statistics hold, so that the obtained AUC equals to the WMW statistic calculated for the function $\predfun_{\mathcal{I}\setminus\{h,i\}}$ on the sample. As an example of extreme inconsistency, we consider what we call a random learning algorithm. This algorithm ignores the training set and randomly infers a prediction function so that they are independent of each other during different rounds of the TLPO cross-validation, which is likely to cause high inconsistency. See supplemental material for more information on tournament inconsistency when a random learning algorithm is used. 

The instability of the learning algorithm depends on the combination of the learning algorithm and the available data. Therefore, the behavior of even the standard learning methods may drift towards one of the previous extreme examples with certain type of data.

The ordinary LPO and TLPO produce exactly the same AUC value, if the tournament in TLPO is consistent. This is obvious as a consistent tournament determines a strict total order on the sample by the edge directions and expressed by the score sequence $0,1,...,\arrowvert\mathcal{I}\arrowvert-1$, and hence $\predfun_{\mathcal{I}\setminus\{h,i\}}(h)<\predfun_{\mathcal{I}\setminus\{h,i\}}(i)$ indicates $S(h)<S(i)$. TLPO in a consistent case then enjoys the same unbiasedness properties as the ordinary LPO. However, the inconsistencies may make the two AUC estimates drift away from each other depending of its severity. 

In the next section, the accuracy of the AUC estimated by LOO, LPO and TLPO in different settings are presented. Moreover, through our experiments we study to which extent inconsistencies in tournaments affect the reliability of TLPO AUC estimates.

\section{Experimental study}
We performed a set of experiments on synthetic and real medical data to evaluate the quality of $\hat{A}_{\textnormal{LOO}}(f)$, $\hat{A}_{\textnormal{LPO}}(f)$ and $\hat{A}_{\textnormal{TLPO}}(f)$, using two different classification methods.  In these experiments we computed the mean and variance of the difference between the estimated and true AUC, over a number of repetitions\cite{braga2004cross,airola2011experimental}. Ideally both quantities should be close to zero. The difference is formally defined as $\Delta \hat{A}_{\textnormal{CV}}(f) = \hat{A}_{\textnormal{CV}}(f) - A(f)$, where $\textnormal{CV}$ refers to one of  $\textnormal{LOO}$,  $\textnormal{LPO}$ or  $\textnormal{TLPO}$. In addition, we carried out an analogous analysis to evaluate the quality of the sensitivity (i.e.TPR) estimated by TLPO at a given specificity (i.e. 1-FPR).

\subsection{Classification methods}
In our experiments the classification methods used were ridge regression and KNN. These methods are widely used learning algorithms due to their simple implementation and high performance. Ridge regression is a representative example of linear and parametric methods, whereas KNN is both a non-linear and non-parametric method. Both methods have the advantage of very fast computation of cross-validation estimates, which makes running the large number of repetitions needed in the simulations computationally feasible. Previously, Airola et al.\cite{airola2011experimental} compared the behavior of ridge regression and support vector machine in cross-validation based AUC estimation, and showed that the methods behaved very similarly.

Ridge regression, also known as regularized least-squares (RLS), is a method that minimizes a penalized version of the least-squared function. \cite{hoerl1970ridge,rifkin2003regularized} This method has a regularization parameter that controls the trade-off between fitting the data and model complexity. The prediction function inferred by this method can be presented in vectorized form as \begin{math} \predfun(\ivector) = \wvector^Tx + b \end{math}, where  \begin{math} \wvector \in \mathbb{R}^{n} \end{math} holds the coefficients of the linear model, \begin{math} \ivector \in \mathbb{R}^{n} \end{math} holds the variables measured from a sample unit for which the prediction is to be made, and $b$ is the intercept. In the simulations that applied ridge regression we used the RLS module from RLScore machine learning library\cite{pahikkala2016rlscore} freely available at \url{https://github.com/aatapa/RLScore}. The regularization parameter was fixed to the value of one, following the same reasons as in Airola et al. (2011).  

KNN is perhaps the simplest and most intuitive of all nonparametric classifiers.\cite{fix1951discriminatory,cover1967nearest} It was originally studied by Fix et al. (1951) and Cover et al. (1967), and it continues to be a popular classifier. Its classification output is based on the majority votes or the average value of the k nearest neighbors. In our experiments a weighted version of KNN was implemented using the Neighbors and the KDtree modules in the scikit-learn library.\cite{scikit-learn} The number of neighbors (k) was set to three, and the output was computed by subtracting the sum of the inverse distance of the negative neighbors from the sum of the inverse distance of the positive neighbors.

\subsection{Synthetic data set}
In the experiments performed on synthetic data, we generated data that reflected the following characteristics: small sample size, class imbalance, low or high dimension, and large number of irrelevant features. The sample size for training the classifier was set to 30 units. The fraction of positives and negatives units in the sample was varied between 10\% and 50\% in steps of 10\%. We considered low-dimensional data with 10 and high-dimensional data with 1000 features. Moreover, both non-signal and signal data were considered.   

In the simulations where no signal occurs in the data, sample units for both positive and negative class were drawn from a standard normal distribution. In the signal data simulations, we considered 10 features with one or four containing signal, and 1000 features with 10 or 50 containing signal. For the discriminating features, the sample units belonging to the positive class were drawn from normal distribution with 0.5 mean and variance one while for the negative class the mean was -0.5.  

The $A(f)$ of a classifier trained on non-signal data is always 0.5, as this classifier can do neither better nor worse than random chance. In contrast, with signal data the $A(f)$ of a classifier trained on a given training set is not known in advance, but it can be estimated from a large test set drawn from the same distribution using equation \eqref{auc}. Therefore, in our signal experiments we used a test set of 10 000 units (5000 positives and 5000 negatives) to compute $A(f)$. Moreover, in order to obtain stable estimates, corresponding mean and variance of $\hat{A}_{\textnormal{LOO}}(f)$, $\hat{A}_{\textnormal{LPO}}(f)$, $\hat{A}_{\textnormal{TLPO}}(f)$ and $A(f)$ were calculated by repeating each simulation 10 000 times. In each repetition a new training data set with same characteristics was sampled.  
\begin{figure}
\centering
\includegraphics[scale=0.47]{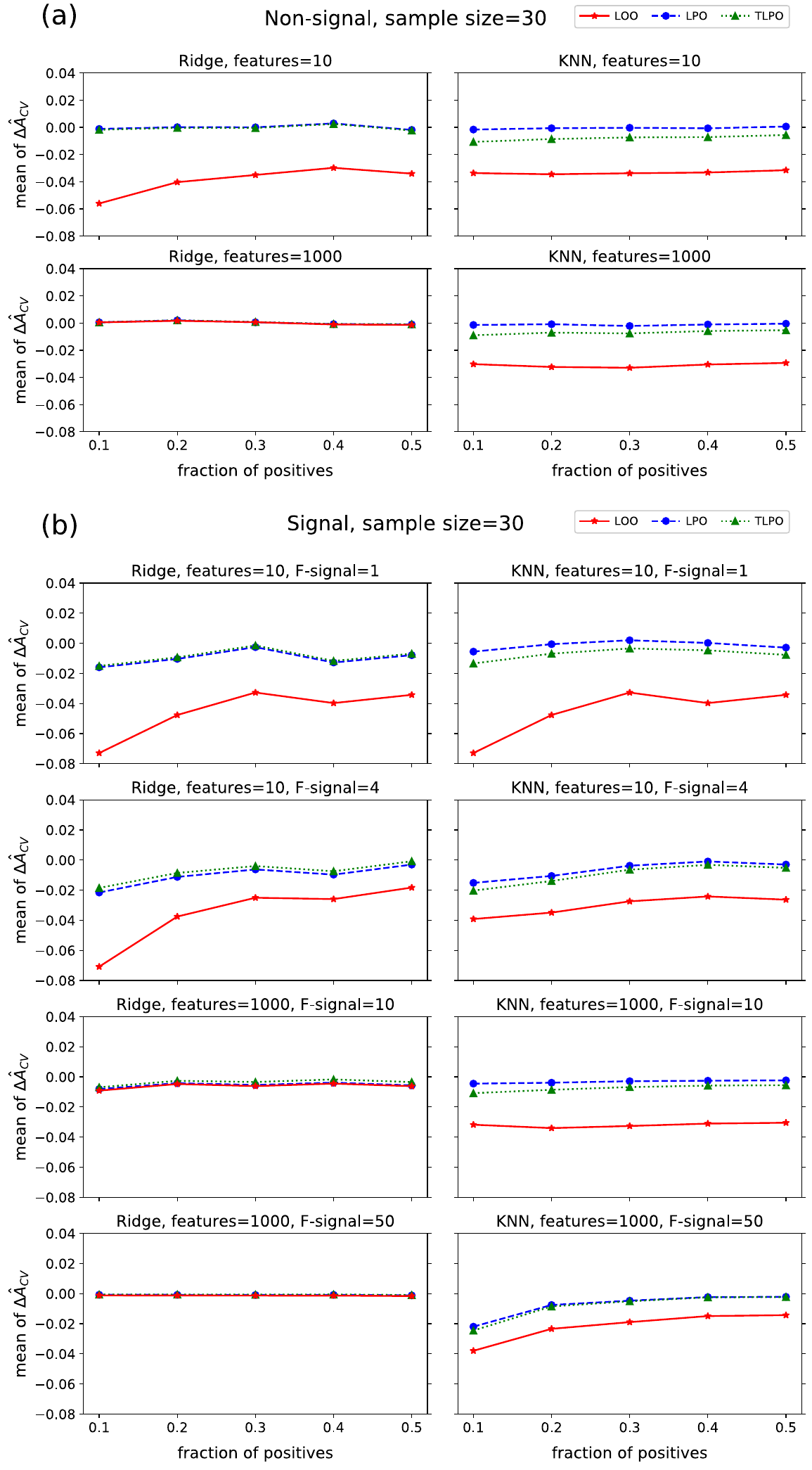}
\caption{\label{Synthetic_DEV} Mean $\Delta \hat{A}_{CV}$ of each cross-validation method over 10 000 repetitions as class fraction balanced on (a) non-signal data (b) signal data. $\Delta \hat{A}_{CV}$: difference between estimated and true AUC; LOO: leave-one-out; LPO: leave-pair-out; TLPO: tournament leave-pair-out; Ridge: ridge regression; KNN: k-nearest neighbors; F-signal: signal features.}
\end{figure}

Figure \ref{Synthetic_DEV}(a) presents mean $\Delta \hat{A}_{CV}$ values of each cross-validation method on non-signal simulations. When using ridge regression, we observe that LPO and TLPO estimates have mean $\Delta \hat{A}_{CV}$ close to zero, and behave similarly regardless of dimensionality or class distribution. LOO estimate compared to LPO and TLPO has a significant negative bias on low-dimensional data. All three estimators behave similarly on high-dimensional data and the negative bias of LOO disappears. The results for LOO agree with the ones reported by Parker et al. (2007), Airola et. al (2011) and Smith et al. (2014). With KNN, LPO mean $\Delta \hat{A}_{CV}$ is close to zero making it a nearly unbiased estimator for this type of classifier. TLPO estimate for KNN shows some negative bias compared to LPO, however, the bias is much smaller than the one shown by LOO. Moreover, supplemental material Figures S2 and S3 show that TLPO with ridge regression had higher consistency in most of our experiments than TLPO with KNN, which may explain the bias for KNN.  

Figure \ref{Synthetic_DEV}(b) displays the mean $\Delta \hat{A}_{CV}$ values of the estimators with signal data. From $\Delta \hat{A}_{CV}$ means we can observe that in this setting all estimators show some bias towards zero, which depends somewhat on the class fraction/sample size and the number of features. This negative bias is inherent to the cross-validation procedure applied to signal data, since the training sets during the cross-validation are slightly smaller than that used to train the final model.

\begin{figure}
\centering
\includegraphics[scale=0.46]{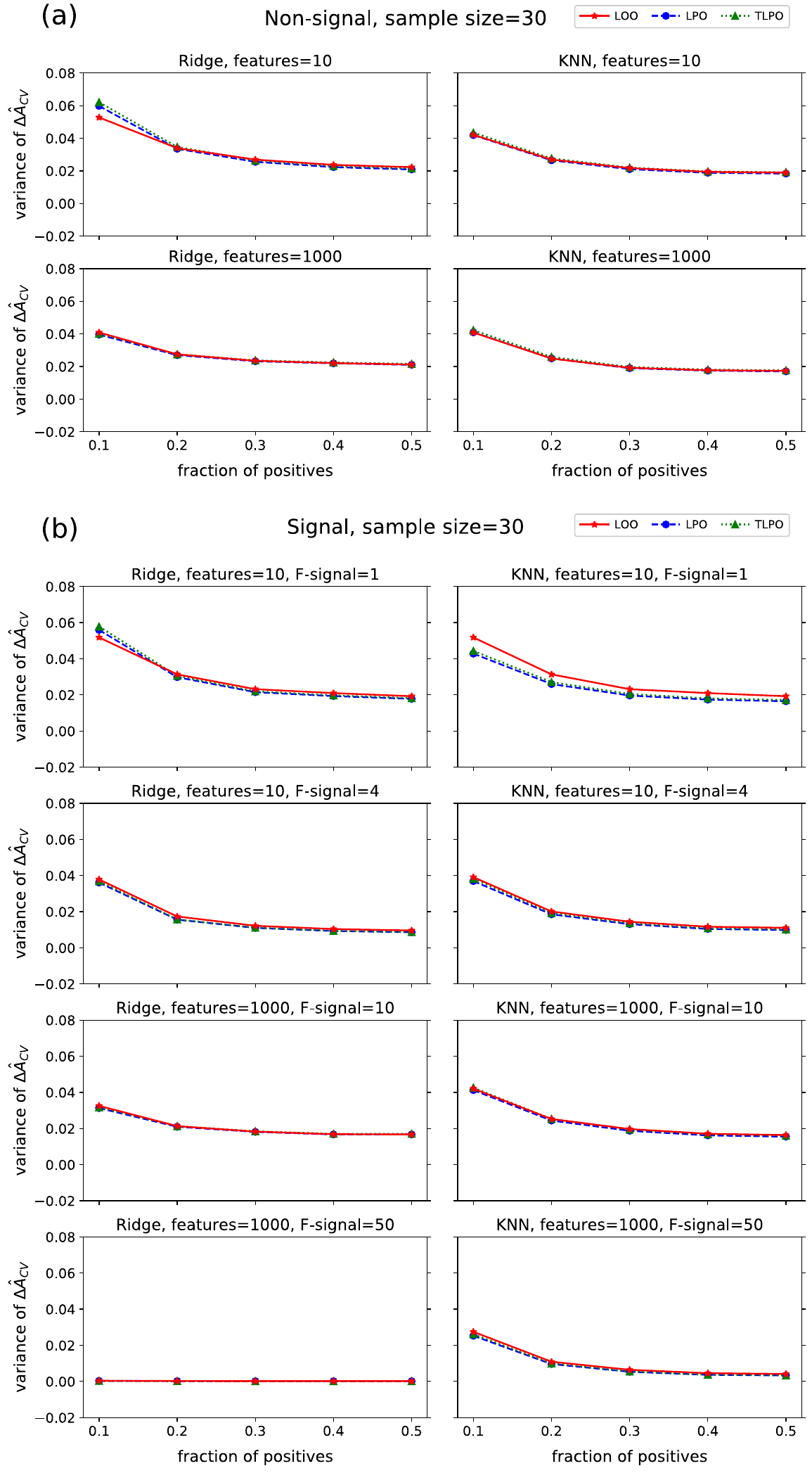}
\caption{\label{Synthetic_VAR} $\Delta \hat{A}_{CV}$ variance of each cross-validation method over 10 000 repetitions as class fraction balanced on (a) non-signal data (b) signal data. $\Delta \hat{A}_{CV}$: difference between estimated and true AUC; LOO: leave-one-out; LPO: leave-pair-out; TLPO: tournament leave-pair-out; Ridge: ridge regression; KNN: k-nearest neighbors; F-signal: signal features.}
\end{figure}

The variances of $\Delta \hat{A}_{CV}$ on non-signal data are presented in Figure \ref{Synthetic_VAR}(a). These results show that $\Delta \hat{A}_{CV}$ variances of all three estimators are higher when there is high class distribution imbalance (only 10\% of positive units in the sample), despite the classification method used. Moreover, there are no notable difference between LOO, LPO and TLPO variances for the more balanced class distributions.    

Figure \ref{Synthetic_VAR}(b) shows the variances of $\Delta \hat{A}_{CV}$ on signal data. Compared to the variances on non-signal data, we observe some similarity when ridge regression is used. However, with KNN and low-dimensional data LOO variances are higher than LPO and TLPO but these differences disappear in high-dimensional data. 

To summarize our results on synthetic data, LPO and TLPO AUC estimates are similar on non-signal and signal data when using ridge regression as classification method. When using KNN, TLPO estimates slightly deviate from LPO estimates, showing some negative bias.  LOO estimates compared to LPO and TLPO presents a much larger negative bias in most of our simulations settings. The variance of the estimates shown in all three cross-validation methods decrease when the class fraction increases. 

\subsection{Real data set}

In addition to synthetic data, we have performed experiments using prostate magnetic resonance imaging (MRI) data to compare
$A(f)$ against $\hat{A}_{LOO}(f)$, $\hat{A}_{LPO}(f)$ and $\hat{A}_{TLPO}(f)$, and to confirm the results obtained on the synthetic data simulations. MRI plays an increasingly important role in the detection and characterization of prostate cancer (PCa) in men with a clinical suspicion of PCa\cite{jambor2015prebiopsy} and those diagnosed with it.\cite{jambor2016prospective} Diffusion weighted imaging (DWI) is the corner stone of prostate MRI. However, validation of DWI post-processing methods is limited due to lack of robust cross validation method.\cite{jambor2016relaxation,merisaari2017fitting,jambor2015evaluation,merisaari2015diffusion} Thus, in this study DWI data of 20 patients with histologically confirmed PCa in the peripheral zone were evaluated. Each patient gave written inform consent and the study was approved the ethical committed of the Turku University Hospital (TYKS) located in Turku, Finland. The DWI data included in this study were part of prior studies focused on development and validation of novel DWI post-processing methods.\cite{jambor2015evaluation,merisaari2015diffusion,toivonen2015mathematical,merisaari2015optimization,perez2016diffusion}

In these experiments, the task was to classify DWI voxels belonging to prostate tumors or non-malignant tissue. The DWI data set consisted of 85 876 voxels (9 268 marked
as cancerous and 76 608 as non-cancerous) obtained from the corresponding parametric maps of 20 patients with PCa. The voxel-wise features were the parameters derived using DWI signal decay modeling: ADCm, ADCk and K as detailed in Toivonen et. al (2015). In addition, Gabor texture was extracted as feature for each parametric map (Gabor-ADCm, Gabor-ADCk, Gabor-K). These six features have shown to have signal in distinguishing tumor voxels from non-tumor voxels. \cite{toivonen2015mathematical,langer2009prostate,ginsburg2011variable} Patient characteristics and image examples are shown in supplemental material (Table 1, Figures S6 and S7).

With this data set we performed experiments varying the class fraction as we did in the synthetic data set simulations. To compute $\hat{A}_{LOO}(f)$, $\hat{A}_{LPO}(f)$ and $\hat{A}_{TLPO}(f)$ we used 30 voxels that were drawn without replacement from the data set. The voxels not drawn were used to calculate $A(f)$. Each experiment was repeated 617 times, as every time a different set of 30 voxels was sampled.

\begin{figure}
\centering
\includegraphics[scale=0.3]{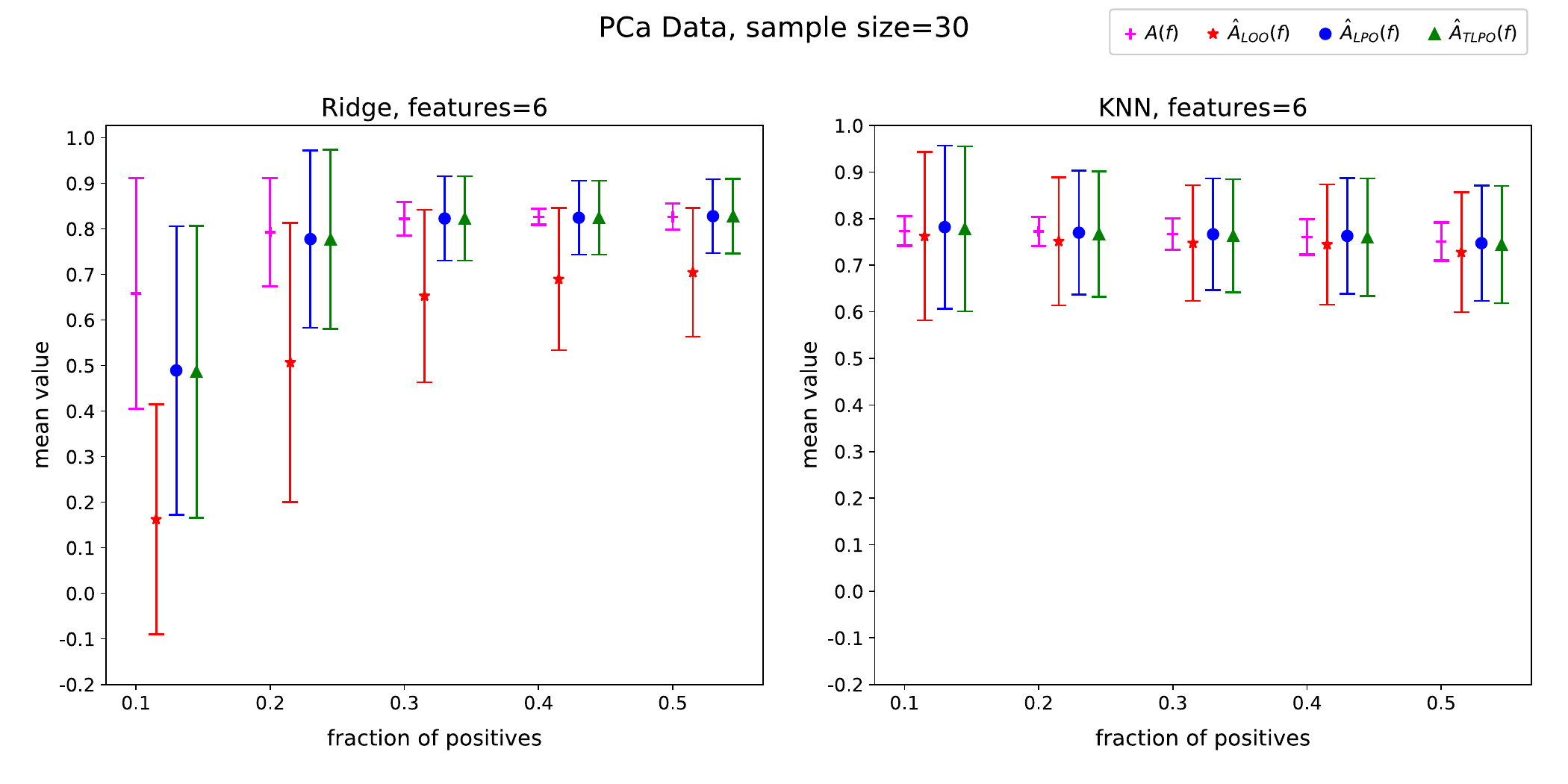}
\caption{\label{PCa_mean_auc} Mean and standard deviation of $A(f)$, $\hat{A}_{LOO}(f)$, $\hat{A}_{LPO}(f)$ and $\hat{A}_{TLPO}(f)$ over 617 repetitions as the class fraction varies. AUC: area under the receiver operating characteristic curve; $A(f)$: true AUC of the classifier; $\hat{A}_{LOO}(f)$: leave-one-out AUC estimate; $\hat{A}_{LPO}(f)$: leave-pair-out AUC estimate; $\hat{A}_{TLPO}(f)$: tournament leave-pair-out AUC estimate; PCa: prostate cancer.}
\end{figure}

\begin{figure}
\centering
\includegraphics[scale=0.3]{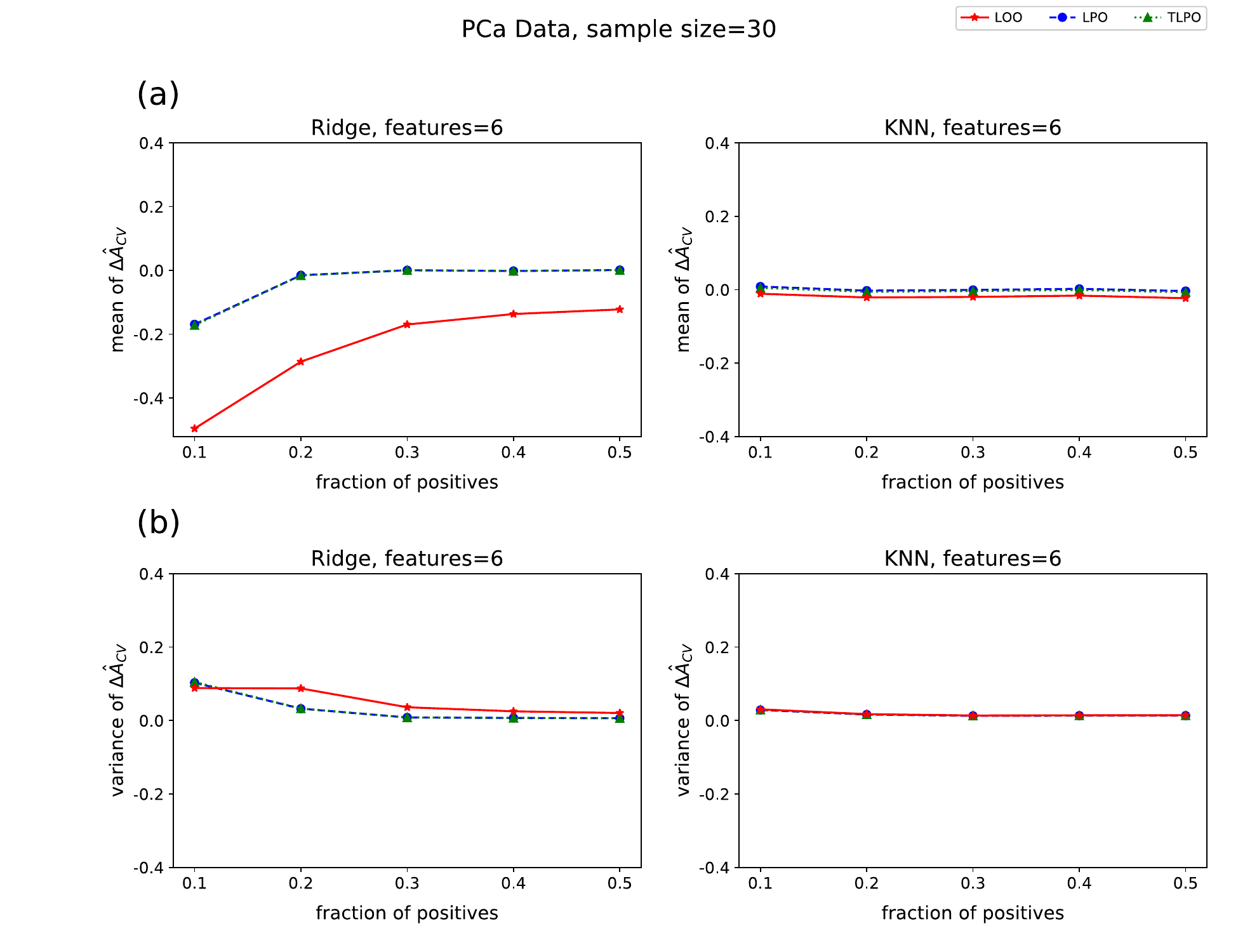}
\caption{\label{PCa_dev_var} (a)  Mean $\Delta \hat{A}_{CV}$ of each cross-validation method on real data  as class fraction varies. (b) $\Delta \hat{A}_{CV}$ variances. $\Delta \hat{A}_{CV}$: difference between estimated and true AUC; LOO: leave-one-out; LPO: leave-pair-out; TLPO: tournament leave-pair-out; Ridge: ridge regression; KNN: k-nearest neighbors; PCa: prostate cancer.}
\end{figure}

The results of these experiments allow us to compare the $A(f)$ to $\hat{A}_{CV}(f)$ of each cross-validation method using real data setting. Figure \ref{PCa_mean_auc} shows the corresponding mean and standard deviation of $A(f)$, $\hat{A}_{LOO}(f)$, $\hat{A}_{LPO}(f)$ and $\hat{A}_{TLPO}(f)$ as the class fraction varies. In these settings we observe that LOO estimates have a strong negative bias when ridge regression is used as classification rule, although, the bias decreases with KNN. In contrast, LPO and TLPO estimates are almost unbiased and only affected by great imbalance among the classes when ridge regression is used. With KNN, LPO and TLPO estimates are unbiased and class imbalance seems not to be affecting the estimates.
The mean $\Delta \hat{A}_{CV}$ values of the estimators are presented in Figure \ref{PCa_dev_var}(a). The results with ridge regression corroborate those obtained on low-dimensional synthetic data. Moreover, the negative bias in TLPO estimates with KNN in the synthetic signal data simulations disappear and there is no significant difference between LPO and TLPO estimates, while LOO estimates still show some negative bias. 
In Figure \ref{PCa_dev_var}(b) the variances of $\Delta \hat{A}_{CV}$ are presented. With ridge regression, we observe high variance in LOO, LPO and TLPO when the class proportion is highly imbalance, but it tends to disappear when the classes are balanced, in same way as in the synthetic data simulation. When KNN is used the variances of all three estimators are close to zero and stable.
\subsection{Sensitivity at a given specificity}

To demonstrate a typical case of ROC analysis made possible by the tournament scores, we considered the estimation of sensitivity of a classifier at given specificity.
We studied the TLPO sensitivity ($\hat{Se}_{\textnormal{TLPO}}$) at a given specificity for all of our experiments using ridge regression and KNN, and compared the estimates with the values obtained from the true ROC curve of the corresponding classifier. The sensitivity values ranged from 0\% to 100\% and specificity from 10\% to 90\% in steps of 10\%. The quality of $\hat{Se}_{\textnormal{TLPO}}$ was measured by the mean and variance of the difference between $\hat{Se}_{\textnormal{TLPO}}$ and true sensitivity ($Se$), defined as  $\Delta \hat{Se}=\hat{Se}_{\textnormal{TLPO}}-Se$, over a number of repetitions. Equations in (\ref{tpr}) were used to compute $Se$, $\hat{Se}_{\textnormal{TLPO}}$ and the specificity (1- FPR). 

\begin{figure}
\centering
\includegraphics[scale=0.32]{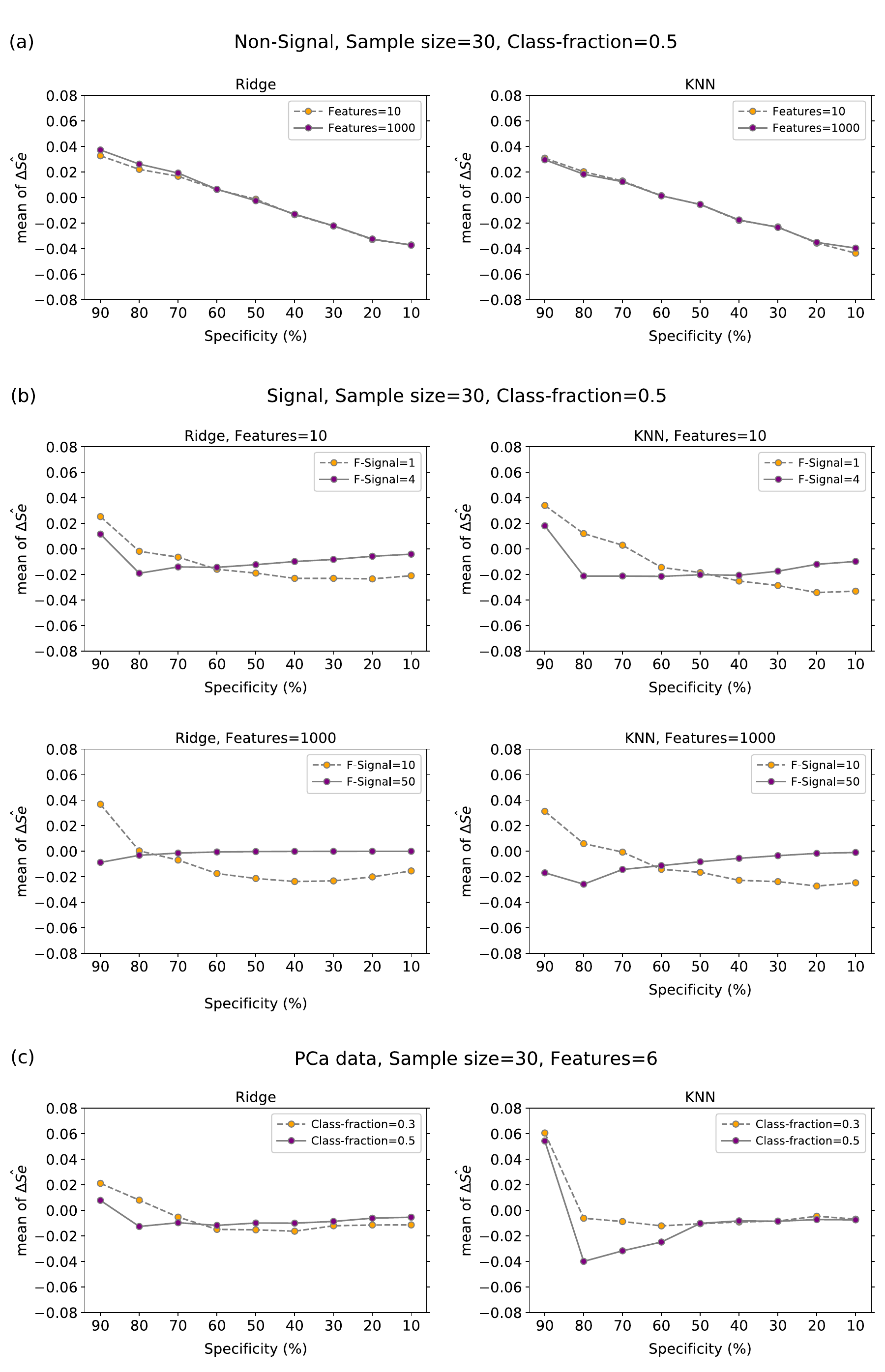}
\caption{\label{mean_se_dev} Mean $\Delta \hat{Se}$ at a given specificity for Ridge and KNN on (a) non-signal data with 10 or 1000 features, (b) signal data with 10 features (one or four have signal) and with 1000 features (10 or 50 have signal) and (c) PCa data with six features for positive class fraction equal to 30\% and 50\%. For (a) and (b) the classes are balanced and the mean value is computed over 10 000 repetitions. In (c) the number of repetitions is 617.  $\Delta \hat{Se}$: TLPO and true sensitivity difference; Ridge: ridge regression; KNN: k-nearest neighbors; Class-fraction: proportion of positive units; F-Signal: number of features with signal; PCa: prostate cancer.}
\end{figure}

\begin{figure}
\centering
\includegraphics[scale=0.32]{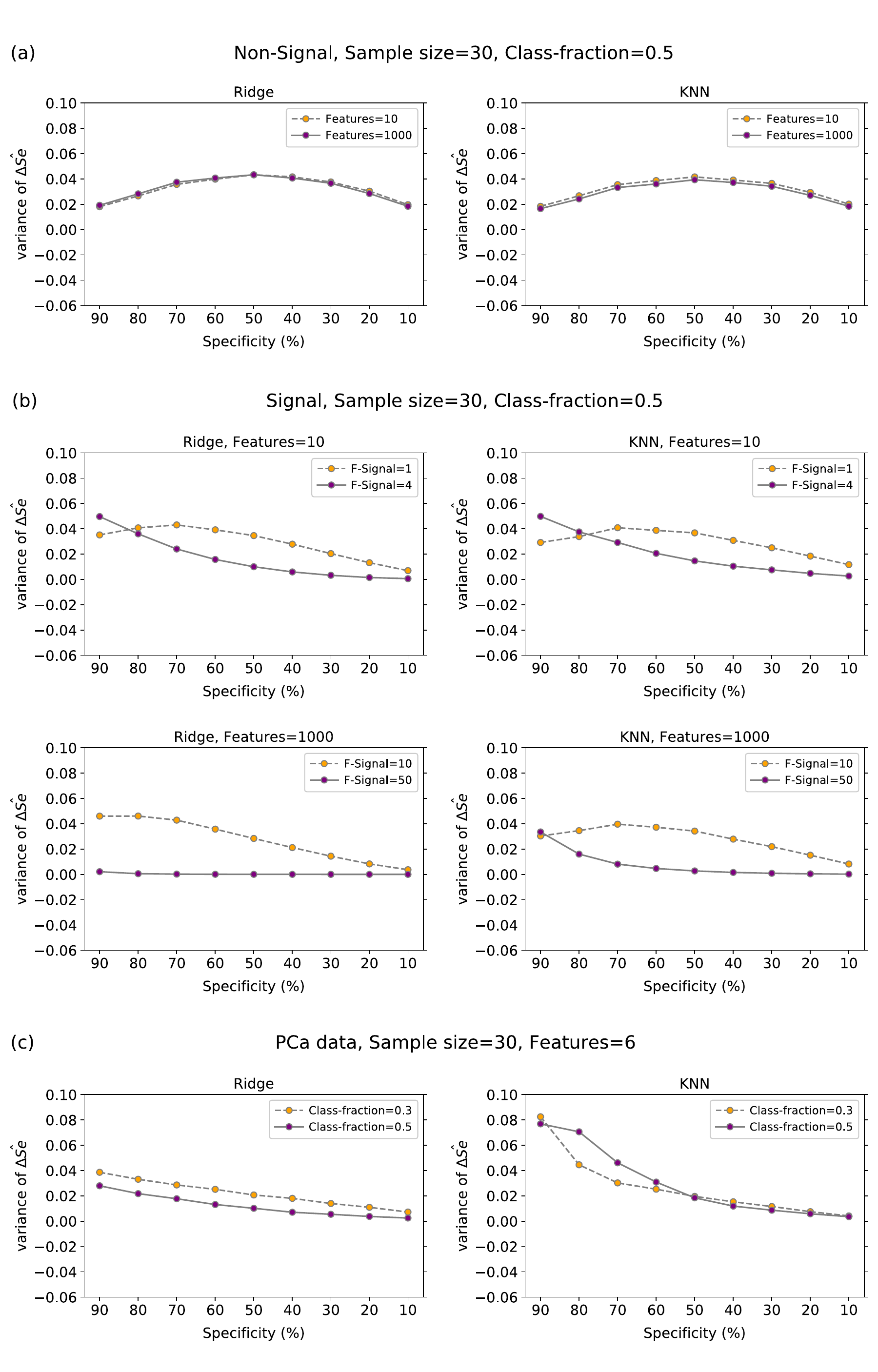}
\caption{\label{var_mean_se_dev} $\Delta \hat{Se}$ variance at a given specificity for Ridge and KNN on (a) non-signal data with 10 or 1000 features, (b) signal data with 10 features (one or four have signal) and with 1000 features (10 or 50 have signal) and (c) PCa data with six features for positive class fraction equal to 30\% and 50\%. For (a) and (b) the classes are balanced and the mean value is computed over 10 000 repetitions. In (c) the number of repetitions is 617. $\Delta \hat{Se}$: TLPO and true sensitivity difference; Ridge: ridge regression; KNN: k-nearest neighbors; Class-fraction: proportion of positive units; F-Signal: number of features with signal; PCa: prostate cancer.}
\end{figure}

Figure \ref{mean_se_dev}(a) presents the mean of $\Delta \hat{Se}$ computed over 10 000 repetitions from non-signal data with 10 and 1000 features and balanced class fraction. In a setting with non-signal data, our results show that there is positive bias for high specificity values and a negative bias for low specificity values. This behavior did not depend on the number of features or classification method used.

The mean of $\Delta \hat{Se}$ computed over 10 000 repetitions from our synthetic signal data is presented in Figure \ref{mean_se_dev}(b). For both ridge regression and KNN, we considered the following settings: low signal in low and high dimension (e.g. 10 features one with signal, 1000 features 10 with signal) and high signal in low and high dimension (e.g. 10 features four with signal, 1000 features 50 with signal). In low signal data, for both classification methods, we observe that mean $\Delta \hat{Se}$ at 90\% specificity is positively biased, but as specificity value decreases the bias also decreases following a similar behavior as in non-signal data. In contrast, when the signal in the data is strong (i.e. the number of features with signal is large) the mean of $\Delta \hat{Se}$ tend to be close to zero when specificity value decreases from 90\% to 10\% with ridge regression while with KNN there is a negative bias in high specificity values.

Figure \ref{mean_se_dev}(c) displays the mean of $\Delta \hat{Se}$ computed over 617 repetitions from our real medical data set. In this case, we analyzed the effect of having balanced classes (i.e fraction of positives = 0.5) against some degree of imbalanced (i.e fraction of positives = 0.3) on mean $\Delta \hat{Se}$. From our results with ridge regression, we notice that for both fractions of positives at 90\% specificity there is some positive bias which decreases and goes close to zero for lower specificity values. With KNN,  at specificity of 90\% for both class fraction there is a positive bias greater that the one observed in ridge regression. However, similar to ridge regression the bias for both class fractions decreases as specificity decreases. Furthermore, in this setting the negative bias for class fraction 0.5 for 80\%, 70\% and 60\% specificity is much greater than the one observed for class fraction 0.3, which is close to zero. 

The variance of $\Delta \hat{Se}$ for all our experiments is presented in Figure \ref{var_mean_se_dev}. In non-signal data, the variance of $\Delta \hat{Se}$ at a given specificity is greater than zero and behaves in a similar manner regardless of the number of features or classification method. On the other hand, the variance of $\Delta \hat{Se}$ in signal data depends somewhat on the number of features with signal and the classification method used. For example, in Figure \ref{var_mean_se_dev}(b) we observe that with ridge regression, high dimension and strong signal the variance of $\Delta \hat{Se}$ is almost zero for all given specificities.

From all results, we observe that sensitivity tends to be more biased near the ends of the ROC curve. This is easily seen from the experiments with the non-signal data, in which the sensitivity is positively biased for large specificity values and negatively  for the small ones. This is a property of ROC curves calculated from a small sample, which can be easily observed from the example case in Figure \ref{TLPO_ROC}. There, sensitivity of the true ROC always approaches zero for 100\% specificity in the limit but it can be considerably larger for finite samples depending on how many positive units are in the top of the ranking that determines the ROC curve. Accordingly, the amount and direction of the bias depends considerably on how much room for variability there is below and above the true ROC curve.

\section{Discussion and future work}

This study proposes TLPO cross-validation for performing ROC analysis and estimating AUC. Our experiments on synthetic data and real data showed that TLPO provides close to unbiased AUC estimates, similarly to the previously proposed LPO method. The advantage of TLPO over LPO is that the former produces also ranking of the data, necessary for computing the ROC curve. Further, our experiments confirmed the substantial negative bias in LOO AUC estimates. Thus, our results suggest that TLPO provides the most reliable cross-validation method for performing ROC curve analysis. This is further backed by an experimental evaluation on computing sensitivity for a given specificity value.

In contrast to using only the positive-negative pairs as when computing AUC with ordinary LPO, we used a complete round robin tournament (or all-play-all tournament) to compute scores for TLPO. This was done for the following reasons. Firstly, it provides simple and convenient consistency analysis tools enabling us to investigate the stability properties of the learning algorithms with respect to LPO by counting the circular triads in the tournament graph. Secondly, the recent theoretical results\cite{balcan2008robust} provide good guarantees for determining a bipartite ranking from a possibly inconsistent tournament. However, in ROC analysis literature, the so-called placement values\cite{Hanley1997Sampling,Pepe2004placement} that are based only on the positive-negative pairs, have been traditionally used for estimating the variance of AUC, comparing two ROC curves or calculating confidence interval for the estimated AUC.\cite{delong1988comparing,qin2008comparison,feng2015comparison} This type of use of placement values together with LPO would be an interesting study of its own, as the effects of the possible inconsistencies in LPO results on these tools is not yet well known. Especially so, since deriving proper confidence intervals for cross-validation is known to be a challenging problem.\cite{bengio2004unbiased}

Future work is required to ascertain to what extent our results generalize to different methods, data distributions and learning methods than those considered in this work. Yet we find it encouraging that similar behavior was observed for the cross-validation methods both on the real and the simulated data. Further, our results about the bias and variance of the LOO and LPO methods are similar to those presented in earlier works,\cite{airola2011experimental,smith2014correcting} where similar results were shown also for larger sample sizes.

Overall, our signal data and real data experiments suggest that if the available data has strong signal TLPO is highly consistent, thus, LPO and TLPO AUC estimates tend to be the same regardless of the classification method. However, it is a good practice to compute both estimates and verify their similarity before performing ROC analysis with TLPO scores. 

\begin{acks}
The authors thank Jussi Toivonen for computing the texture features for the medical data set, and the anonymous reviewers for constructive comments, especially regarding the connection of tournament scores to placement values in ROC analysis.
\end{acks}

\begin{dci}
The author(s) declare(s) that there is no conflict of interest.
\end{dci}

\begin{funding}
This work was supported by Turku University Foundation (Pentti ja Tyyni Ekbomin Rahasto grant 13757) and by the Academy of Finland (grants 289903, 311273).
\end{funding}

\bibliographystyle{SageV}
\bibliography{myBibliography}        

\begin{sm}
\setcounter{figure}{0}
\makeatletter 
\renewcommand{\thefigure}{S\arabic{figure}}
\\
\textbf{1. Bootstrapping for adjusting AUC estimate}. To study the quality of the bootstrap adjusted AUC estimate described by Harrell et al.\citep{harrell1996multivariable}, we followed the bootstrap procedure and performed experiments on non-signal data sets of size 30 using 10 and 1000 features with ridge regression and KNN as classification methods.The number of bootstrap sampled with replacement from the original data set was 200.  Each experiment was repeated 10 000 times to compute the mean and variance of the difference between the adjusted AUC estimate and the true AUC, formally defined as $\Delta \hat{A} = \hat{A} - 0.5$, where 0.5 is the true AUC for the data with no signal. Figure \ref{Non_Signal_cv_boots}(a) presents the mean of $\Delta \hat{A}$ for the bootstrap adjusted AUC estimate (BOOTS) and for each of the cross-validation methods (i.e. LOO, LPO and TLPO) estimates as the positive class fraction varies. In all our experiments with non-signal data BOOTS shows an optimistic bias whereas the bias of the cross-validation methods is close to zero or is pessimistic as in the case of LOO. The variance of $\Delta \hat{A}$ for each of our experiments is shown in Figure \ref{Non_Signal_cv_boots}(b), from these results we observe that  BOOTS has variance close to zero and lower than the variance of the cross-validations methods. The high AUC value obtained by bootstrap even on data with no signal indicates that the experimenter can not distinguish between the cases in which the learning algorithm has overfit to the data or it has actually learned a useful classifier. Due to its low variance bootstrap can still be a valuable method when using classical statistical methods and low-dimensional data. It can become quite biased when working with more expressive methods that can always fit themselves to their training data to such extent that they can predict it perfectly. This is true for example of the KNN method, modern machine learning methods such as (deep) neural networks, and even linear models when having more features than samples.\\
\begin{figure}
\centering
\includegraphics[scale=0.38]{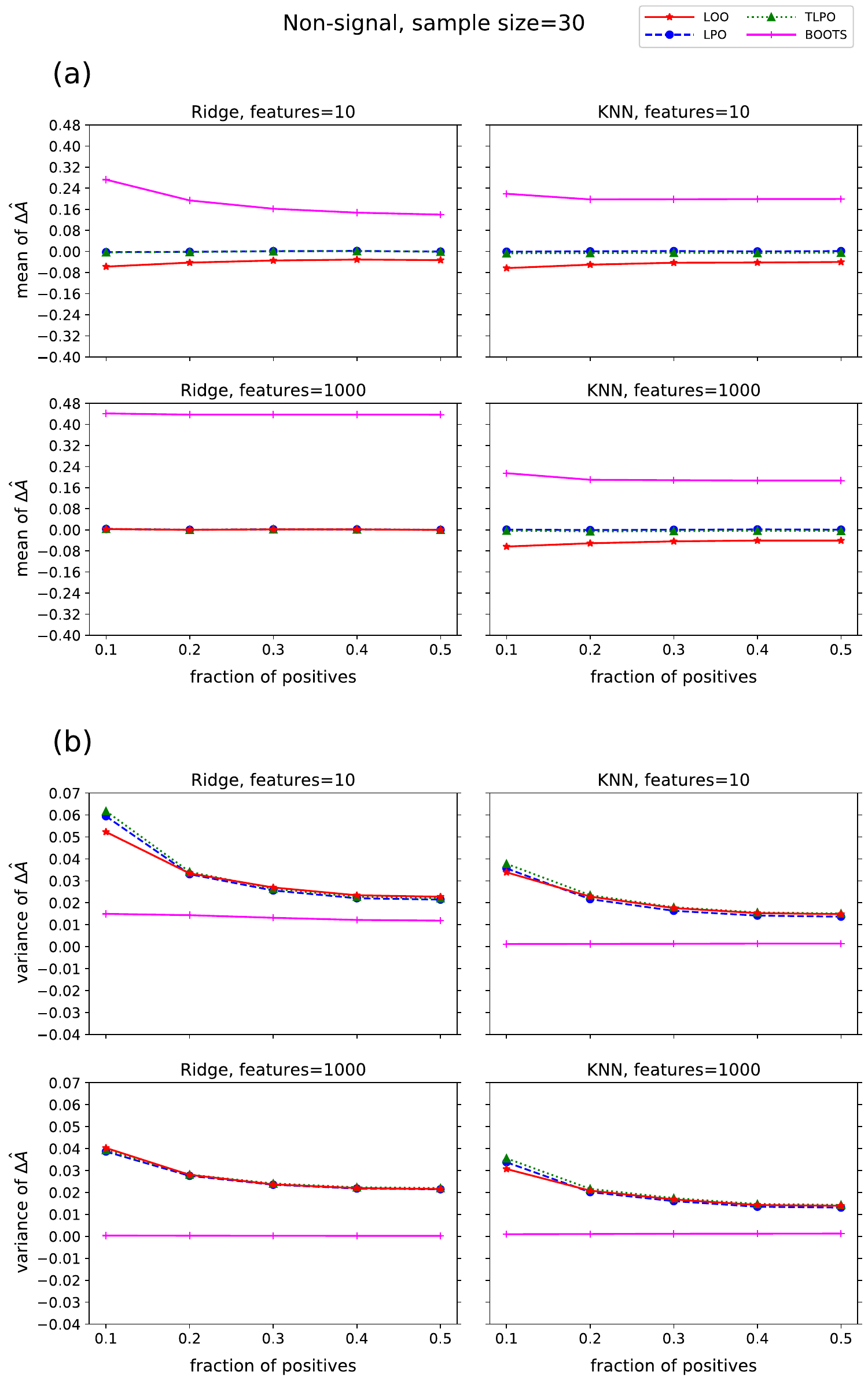}
\caption{\label{Non_Signal_cv_boots}(a) Mean $\Delta \hat{A}_{CV}$ and (b) $\Delta \hat{A}_{CV}$ variance of LOO, LPO, TLPO, and BOOTS over 10 000 repetitions for all our experiments in non-signal data as class-fraction balanced. $\Delta \hat{A}_{CV}$: difference between estimated and true AUC; LOO: leave-one-out; LPO: leave-pair-out; TLPO: tournament leave-pair-out; BOOTS: method for adjusting optimistic bias present in the resubstitution estimate using 200 bootstrap samples; Ridge: ridge regression; KNN: k-nearest neighbors.}
\end{figure}\\
\textbf{2. Level of inconsistency}. In a tournament with $m$ sample units the level of inconsistency can be measured, as explained in \cite{kendall1940method,harary1966theory,gass1998tournaments}, by counting the number of circular triads ($c$) in the corresponding graph
\begin{align*}
c = m(m-1)(2m-1)/12 - \frac{1}{2}\sum_{i=1}^m S(i)^2\;,
\end{align*}
where $S(i)$ is the score for unit $i$. The maximum number of circular triads in a tournament graph is define by  
\begin{align*}
c_{max}(m) =\begin{cases}\frac{m^3-m}{24} & when \ m \ is \ odd, and\\
\frac{m^3-4m}{24} & when \ m \ is \ even.\end{cases} 
\end{align*}
Kendall and Babington Smith \cite{kendall1940method} coefficient of consistency ($\xi$) is then

\begin{align*}
\xi = 1-\frac{c}{c_{max}(m)} = \begin{cases}1-\frac{24c}{m^3-m} & when \ m \ is \ odd, and\\1-\frac{24c}{m^3-4m} & when \ m \ is \ even.\end{cases} 
\end{align*}\\
\textbf{3. Tournament inconsistency when a random learning algorithm is used for training}. A random learning algorithm ignores the training set and infers a prediction function, say with values in range $[-1,1]$, that is randomly drawn from an uniform distribution over the set of functions $[-1,1]^\mathbb{N}$. When a random learning algorithm is used with TLPO, the prediction functions inferred during different rounds of the cross-validation are independent of each other, and hence the number of cycles follow the distribution of cycle amounts of a tournament graph with random edge directions. In the case of TLPO with a random learning algorithm and a sample size of 30, the expected value of $\xi$ is 0.1 indicating a high level of inconsistency.\\ \\
\textbf{4. Coefficient of consistency $\xi$ of TLPO in the experiments with synthetic data and real medical data}. Mean of $\xi$ for each of the experiment in our study are presented in Figures \ref{No_Signal_CC}, \ref{Signal_CC}, \ref{Pca_CC}. In most of our synthetic experiments, TLPO with KNN is less consistent than Ridge regression. Figure  \ref{No_Signal_CC} and \ref{Signal_CC} show that with KNN $\xi$ is 0.97 or below, while Ridge regression $\xi$ is 0.96 or above. This may be the reason of the small negative bias observed on TLPO AUC estimate with KNN. \\
In our experiments with the real medical data set, TLPO with both learning algorithms, Ridge regression and KNN, has $\xi$ above 0.97 (figure \ref{Pca_CC}). In these experiments, the small negative bias previously observed in TLPO AUC estimate with KNN disappears.
\begin{figure}
\centering
\includegraphics[scale=0.35]{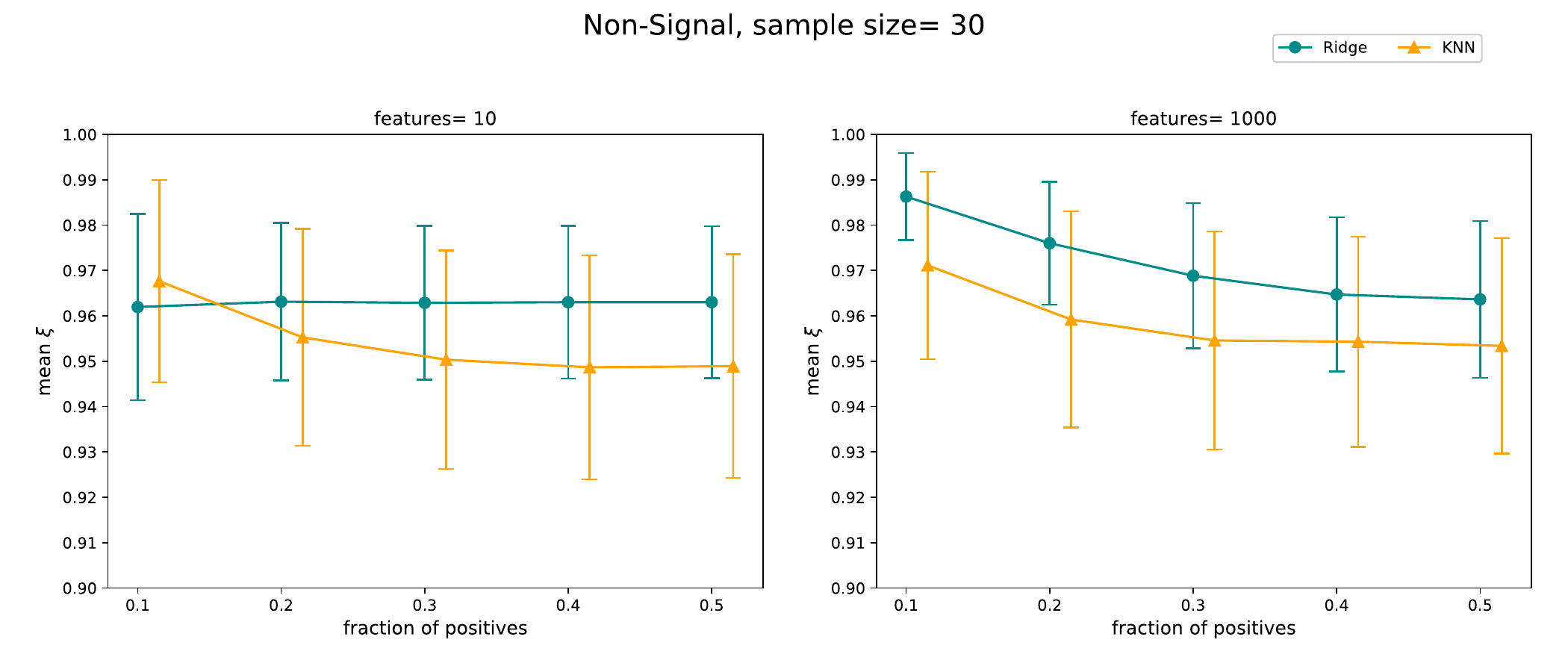}
\caption{\label{No_Signal_CC} Mean of the coefficient of consistency $\xi$ for TLPO on non-signal data, using ridge regression (Ridge) and K-nearest neighbors (KNN) as learning algorithms and varying the fraction of positives. Each experiment was repeated 10 000 times.}
\end{figure}
\begin{figure}
\centering
\includegraphics[scale=0.34]{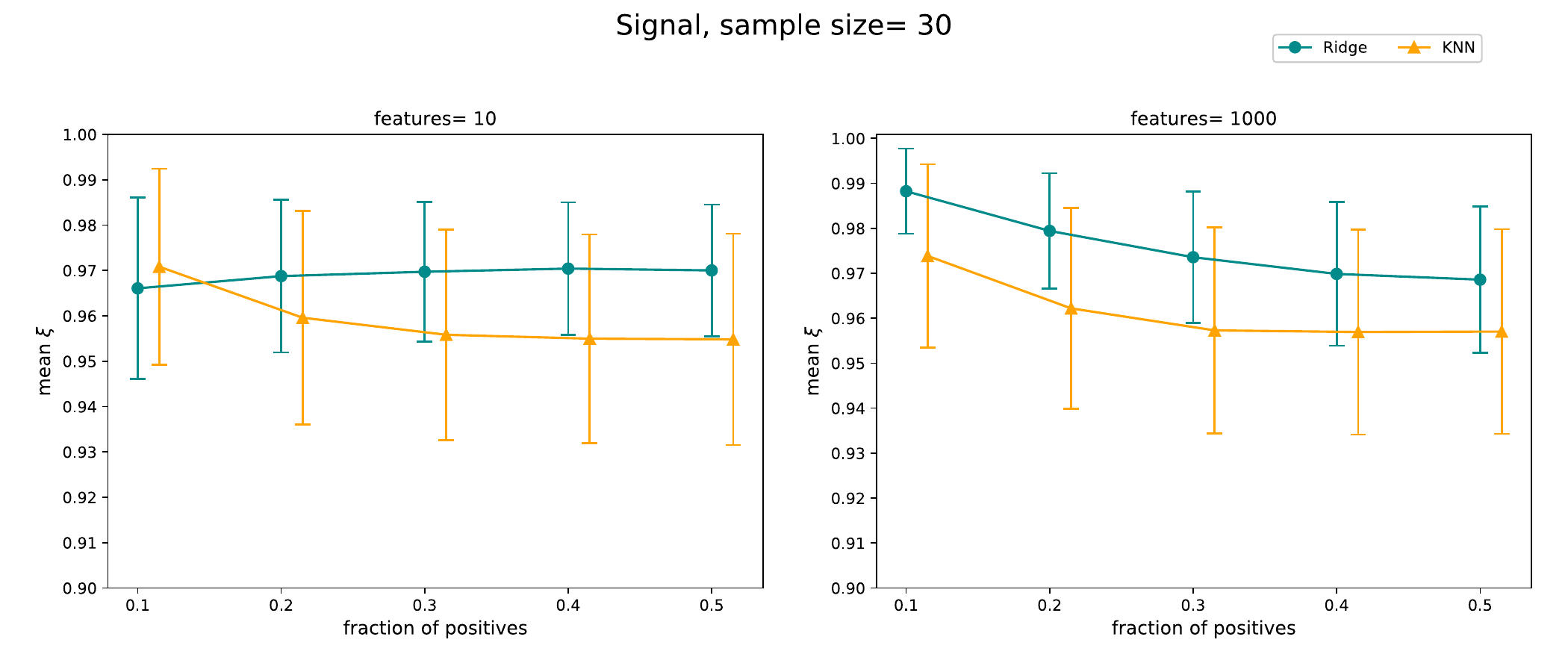}
\caption{\label{Signal_CC} Mean of the coefficient of consistency $\xi$ for TLPO on signal data, using ridge regression (Ridge) and K-nearest neighbors (KNN) as learning algorithms and varying the fraction of positives. Each experiment was repeated 10 000 times.}
\end{figure}
\begin{figure}
\centering
\includegraphics[scale=0.38]{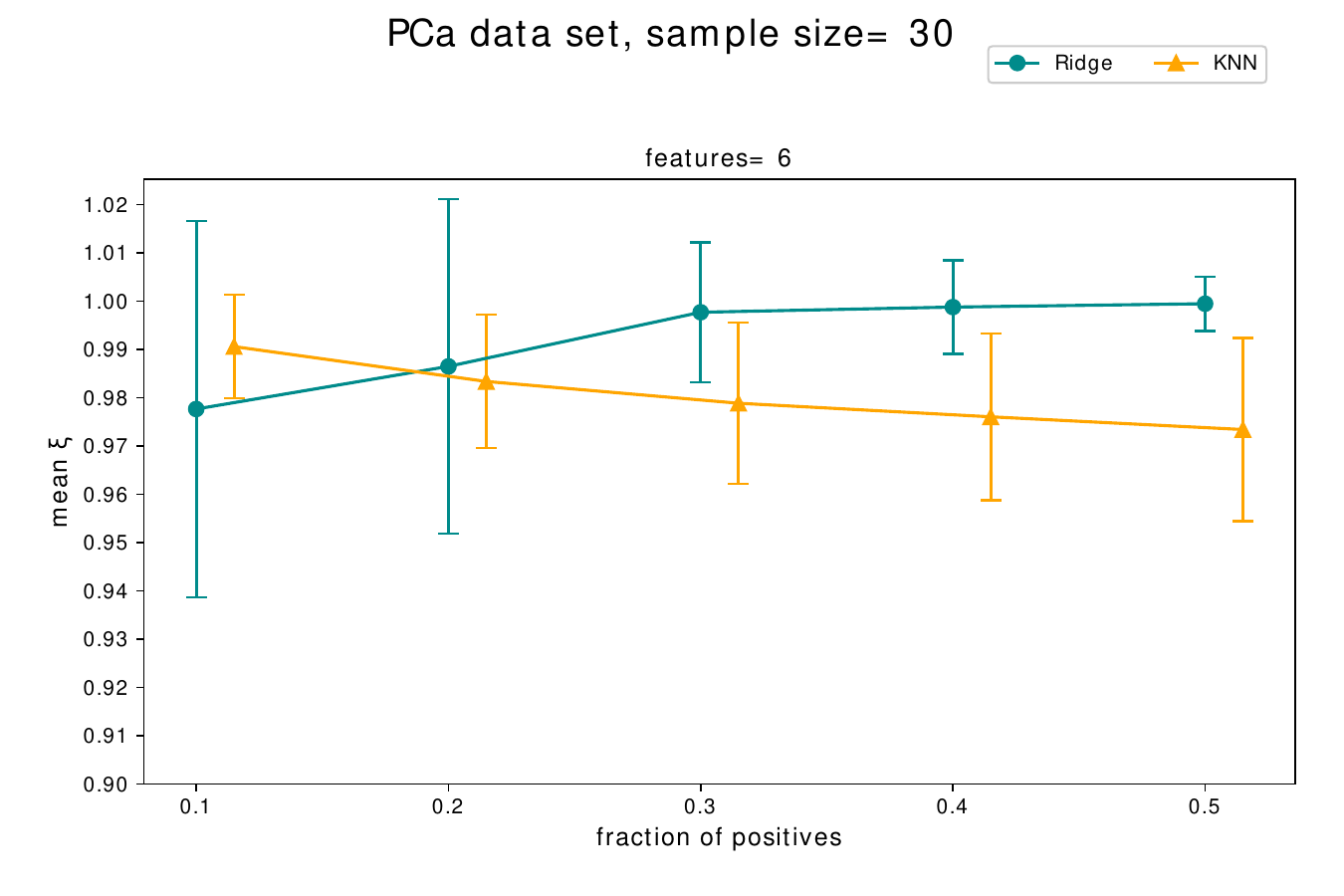}
\caption{\label{Pca_CC} Mean of the coefficient of consistency $\xi$ for TLPO on real medical prostate cancer (PCa) data set, using ridge regression (Ridge) and K-nearest neighbors (KNN) as learning algorithms and varying the fraction of positives. Each experiment was repeated 617 times.}
\end{figure}
\begin{table}[htp]
\small\sf\centering
\caption{Characteristics of the patients included in the real medical data set. The tumors are located in the prostate peripheral zone. PSA: prostate specific antigen.}
\begin{tabular}{clll}
\toprule
Patient no.&Age (years)&PSA(ng/ml)&Gleason score\\
\midrule
1&67&11&3+4\\
2&66&9.3&4+3\\
3&68&30.0&5+4\\
4&66&15.0&4+5\\
5&67&12.0&3+4\\
6&68&3.9&3+4\\
7&60&28.0&4+3+5\\
8&62&7.7&3+4\\
9&67&5.1&4+4\\
10&65&5.7&3+4\\
11&65&9.9&3+4\\
12&70&12.0&3+4+5\\
13&62&4.1&4+5\\
14&67&4.6&4+5\\
15&67&8.3&4+3+5\\
16&66&6.6&4+3\\
17&45&12&3+4\\
18&60&8.6&4+5\\
19&65&4.5&3+4\\
20&68&3.2&3+4\\
\bottomrule
\end{tabular}
\end{table}
\begin{figure}
\centering
\includegraphics[scale=0.9]{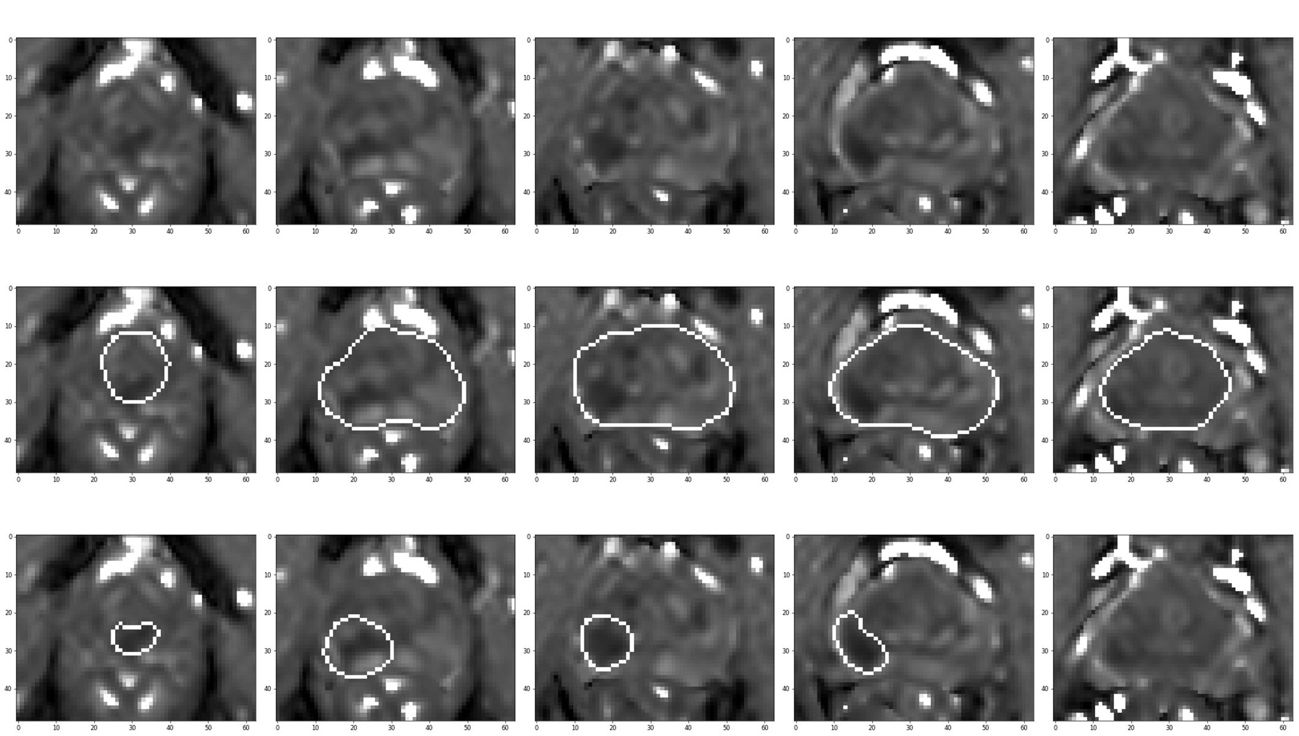}
\caption{\label{ADC_pat_1} Images from patient no. 1. First Row: ADCm map of prostate. Second Row: prostate delineated. Third Row: tumor delineated}
\end{figure}
\begin{figure}
\centering
\includegraphics[scale=0.9]{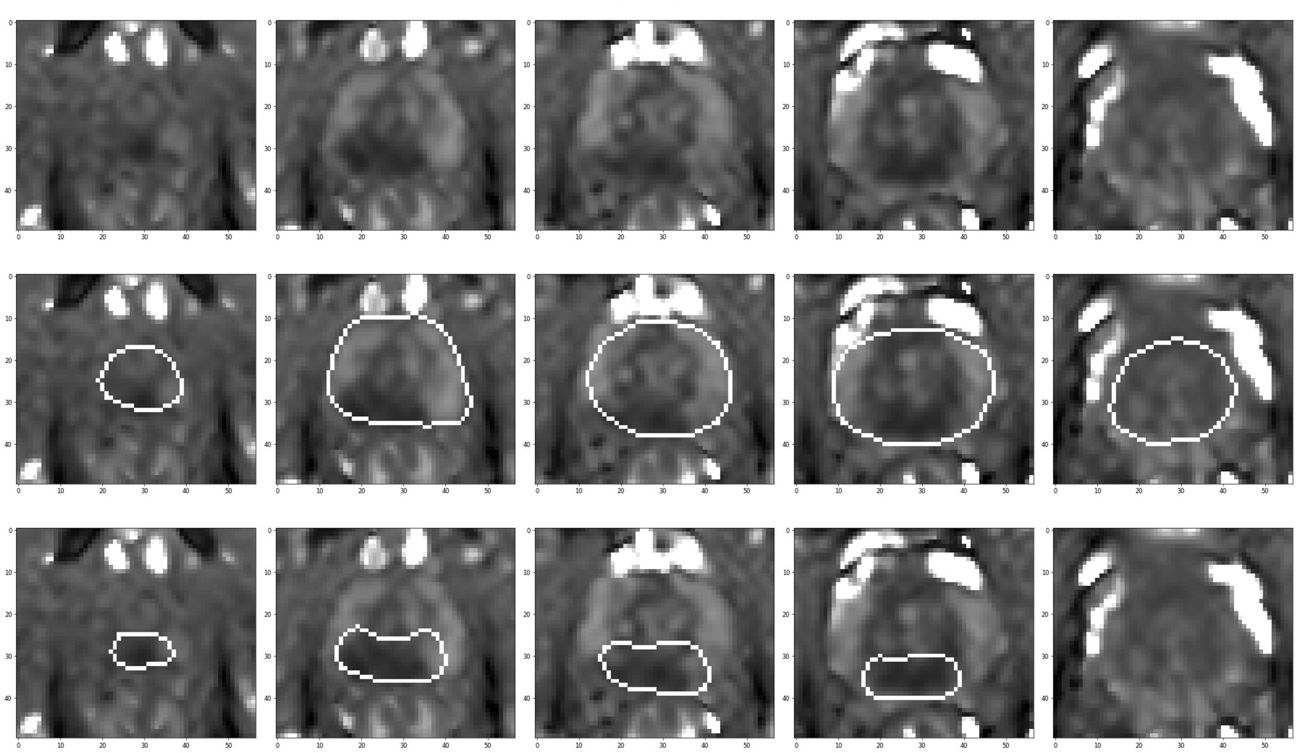}
\caption{\label{ADC_pat_4} Images from patient no. 4. First Row: ADCm map of prostate. Second Row: prostate delineated. Third Row: tumor delineated}
\end{figure}
\end{sm}
\end{document}